\documentclass[a4paper,11pt]{article}
\usepackage{pos}

\usepackage{amsmath}

\usepackage{amssymb}
\usepackage{graphicx}
\usepackage{booktabs}
\usepackage{xcolor}

\title{Towards an Autonomous AI Steward for Large-Scale Physics
Experiments: Integrating Causal Diagnostics, RAG, and Conformal
Prediction}
\ShortTitle{Towards an Autonomous AI Steward for Physics Experiments}

\author*[a]{Oliver Noll}
\author[b]{Nicholas Tan Jerome}
\author[a]{Alexander Lehnen}
\author[b]{Frank Simon}
\author[a]{Frank Maas}

\affiliation[a]{Helmholtz Institute Mainz (HIM),\\
  GSI Helmholtzzentrum f\"ur Schwerionenforschung,\\
  Mainz, Germany}

\affiliation[b]{Institute for Data Processing and Electronics (IPE),\\
  Karlsruhe Institute of Technology (KIT),\\
  Karlsruhe, Germany}

\emailAdd{o.noll@gsi.de}
\emailAdd{nicholas.jerome@kit.edu}

\abstract{
Long-term particle physics experiments face critical operational
challenges: monitoring non-stationary machine data streams and
preserving tacit expert knowledge. We present the ``Helmholtz
Co-Pilot,'' a hybrid Multi-Agent System (MAS) acting as an autonomous
experiment steward. The system can be triggered interactively by human
operators or autonomously by an asynchronous causal diagnostics engine
that runs independently of the MAS reasoning core. Continuously
monitoring live hardware streams such as data acquisition and slow
control, this engine performs dynamic causal discovery to trace
cascading alarms back to their root causes and applies adaptive model
selection to keep forecasts reliable across shifting machine states;
upon detecting an anomaly, it hands a structured diagnostic report to
the MAS. The reasoning layer itself relies on a high-throughput C++
core and a Python edge architecture orchestrating a directed acyclic
graph (DAG) of specialized ReAct workers. It integrates a
Retrieval-Augmented Generation (RAG) backbone for institutional memory
and leverages conformal prediction to bound its evidence-based
recommendations with mathematical coverage guarantees. This extended
abstract outlines the architecture, the integration of
software-enforced Human-in-the-Loop (HitL) guardrails, and ongoing
shadow-testing within the KATRIN experiment.
}

\FullConference{2nd International Conference on AI for Science  (AI4Sci2026)\\
5-9 October 2026\\
Mainz, Germany\\}

\begin{document}
\maketitle

% ---------------------------------------------------------------------
\section{Introduction}
\label{sec:intro}

Decadal physics experiments, like KATRIN~\cite{ref_katrin} and
P2~\cite{ref_p2}, rely on stable core teams alongside rotating student
and doctoral operators. Despite extensive resources for wikis and
threshold-based alarms, onboarding personnel, preserving tacit
knowledge, and isolating root causes during cascade alerts remain key
challenges. To augment existing infrastructures, we introduce the
``Helmholtz Co-Pilot,'' an autonomous digital colleague acting as a
bridge between human intuition and machine telemetry. The system is
designed to be triggered by two distinct inputs: interactively by a
shift operator querying the system, or autonomously by an asynchronous
causal diagnostics engine detecting hardware anomalies. Regardless of
the trigger, the request is processed by a unified, intelligent
orchestration layer that bridges human institutional memory and
data-driven hardware diagnostics.

% ---------------------------------------------------------------------
\section{Core Architecture: The Hybrid MAS Engine}
\label{sec:architecture}

\begin{figure}[t]
  \centering
  \includegraphics[width=\textwidth]{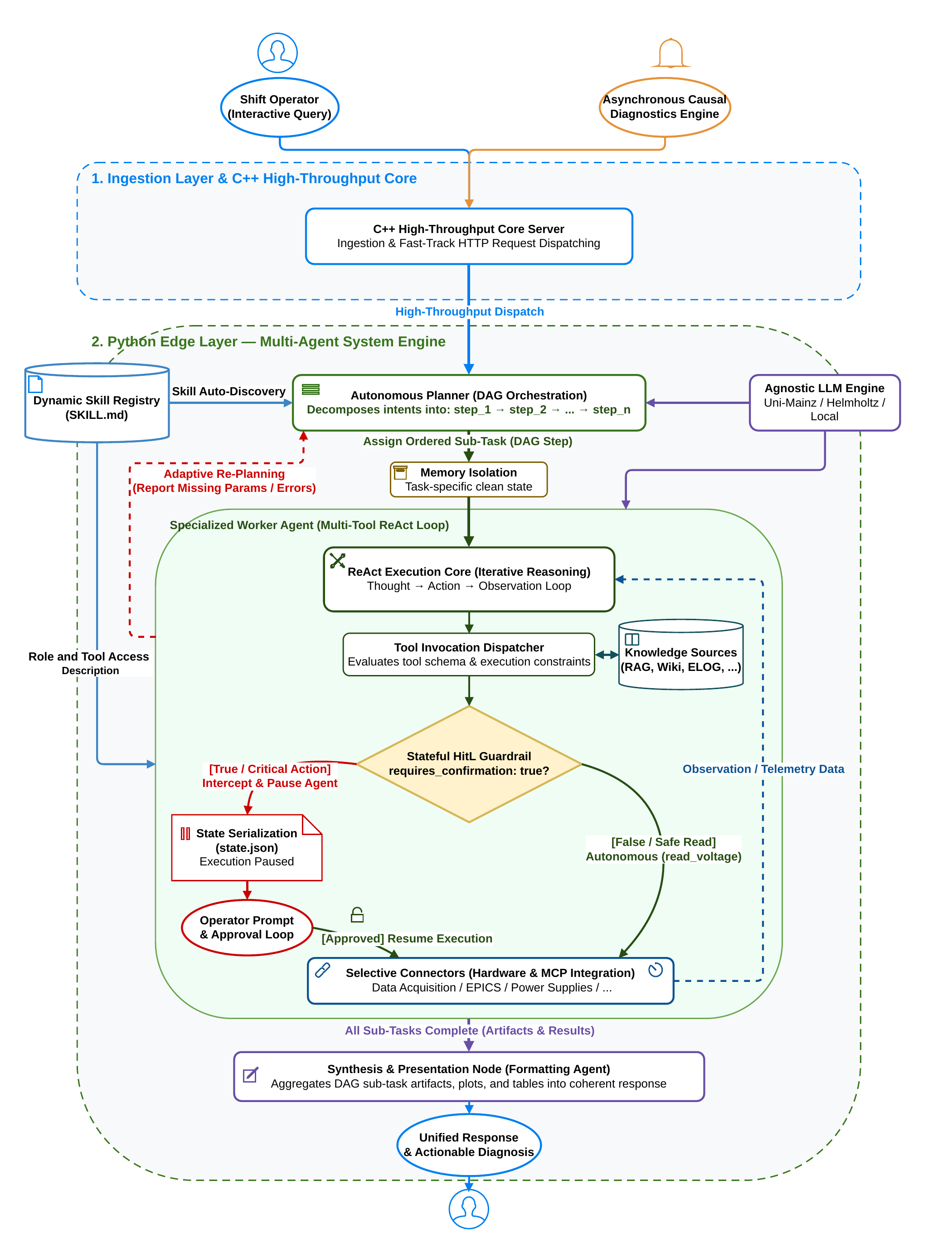}
  \caption{Core architecture of the Helmholtz Co-Pilot. Input streams
  are ingested via a C++ Core and dispatched to the Python Edge Layer.
  An Autonomous Planner generates a DAG of sub-tasks executed by ReAct
  Workers under strict HitL guardrails.}
  \label{fig:architecture}
\end{figure}

The Co-Pilot relies on an adaptive Multi-Agent System
(MAS)~\cite{ref_mas} built upon a hybrid architecture designed for both
high-throughput telemetry and complex natural language reasoning
(Fig.~\ref{fig:architecture}).

\noindent\textbf{1. Ingestion Layer \& C++ Core.}
To handle the scale of physical experiments, the system utilizes a
high-throughput C++ core server. It ingests fast-track HTTP requests
from both interactive operator queries and autonomous diagnostic
engines, dispatching them securely to the agentic reasoning
environment.

\noindent\textbf{2. Python Edge Layer \& DAG Orchestration.}
The reasoning heavily relies on a Python edge layer powered by an
agnostic LLM engine. While capable of integrating arbitrary commercial
providers, the architecture prioritizes sovereign institutional
endpoints (e.g., university clusters, research centers, and local
fine-tuned models). An \textit{Autonomous Planner} utilizes skill
auto-discovery via a dynamic skill registry (leveraging
\texttt{SKILL.md}, an emerging standard combining Markdown and YAML) to
evaluate available capabilities. It then decomposes complex intents
into a Directed Acyclic Graph (DAG) of ordered sub-tasks, ensuring
sequential, deterministic execution paths.

\noindent\textbf{3. ReAct Workers, HitL Guardrails \& Synthesis.}
Each DAG step is assigned to a \textit{Specialized Worker Agent}
utilizing a multi-tool ReAct (Reason and Act) loop. To prevent context
poisoning, workers operate in strict memory isolation with
task-specific clean states. Missing parameters or execution errors
trigger an adaptive re-planning loop back to the orchestrator.
Crucially, the execution core evaluates tool schemas through a stateful
Human-in-the-Loop (HitL) guardrail. While safe read operations proceed
autonomously, critical actions (flagged via
\texttt{requires\_confirmation: true}) intercept the agent, serialize
its state, and pause execution until the shift operator approves the
action via a prompt loop. Finally, a \textit{Formatting Agent} operates
as a standard ReAct worker within the agent pool. Invoked by the
planner at the end of a DAG, or during intermediate steps if required,
it aggregates all generated artifacts, retrieved knowledge, and
telemetry data into a unified, actionable diagnosis.

% ---------------------------------------------------------------------
\section{The Two Operational Pillars}
\label{sec:pillars}

Importantly, both human queries and autonomous machine alerts can
trigger dynamic workflows across the system. The Co-Pilot
operationalizes these workflows through two foundational pillars,
seamlessly blending them to assist the operator.

\noindent\textbf{Pillar 1: Institutional Memory (Human Side).}
This pillar utilizes knowledge sources (DokuWiki, ELOG, databases,
\ldots) via a Retrieval-Augmented Generation (RAG) backbone. To
evaluate this layer without hallucinating constants, we introduce
\textit{PhysRAG}, a domain-specific benchmark built from expert
elicitation. Furthermore, the system proactively triggers an interview
mode upon detecting knowledge gaps, writing heuristic operator insights
directly back to the vector stores.

\noindent\textbf{Pillar 2: Hardware Diagnostics (Machine Side).}
Operating independently from the MAS reasoning core, an asynchronous
causal diagnostics engine continuously monitors live hardware streams
(e.g., data acquisition and slow control). It performs dynamic causal
discovery to trace cascading alarms back to their root causes and
utilizes adaptive model selection for reliable forecasting across
shifting machine states. When a machine anomaly occurs, this engine
autonomously triggers the MAS with a diagnostic report. The MAS
immediately processes this telemetry, querying the knowledge base
(Pillar 1) for mitigation procedures. Leveraging conformal
prediction~\cite{ref_cp}, the MAS translates these insights into
evidence-based decisions. It presents the operator with a highly
precise, synthesized state overview bounded by mathematical coverage
guarantees that quantify the exact probability of a specific root cause
alongside actionable troubleshooting steps.

% ---------------------------------------------------------------------
\section{Conclusion and Ongoing Work}
\label{sec:conclusion}

The Helmholtz Co-Pilot provides a scalable blueprint for trustworthy AI
in fundamental physics. The MAS core is currently transitioning from a
prototype into a full-fledged framework featuring an optimized user
interface, robust terminal integrations, and streamlined deployment via
Docker containers. The source code is maintained in a dedicated Git
repository to promote open science and accessibility. Ongoing efforts
focus on shadow-testing within KATRIN~\cite{ref_katrin} to evaluate
real-world performance safely.

% ---------------------------------------------------------------------
\section*{Acknowledgements}

Supported by the Helmholtz Association and participating centers
HIM/GSI and IPE/KIT.

% ---------------------------------------------------------------------


\begin{thebibliography}{4}

\bibitem{ref_katrin}
  KATRIN Collaboration: The design, construction, and commissioning of
  KATRIN. JINST \textbf{16}(08), T08015 (2021)

\bibitem{ref_p2}
  Becker, D., et al.: The P2 experiment -- A future high-precision
  measurement of the electroweak mixing angle at low momentum transfer.
  Eur. Phys. J. A \textbf{54}, 208 (2018)

\bibitem{ref_mas}
  Chen, S., et al.: A Survey on LLM-based Multi-Agent System.
  arXiv:2412.17481 (2024)

\bibitem{ref_cp}
  Angelopoulos, A. N., Bates, S.: A Gentle Introduction to Conformal
  Prediction and Distribution-Free Uncertainty Quantification.
  arXiv:2107.07511 (2021)

\end{thebibliography}
\end{document}